\documentclass[12pt,a4paper,twoside,article]{memoir}
\usepackage[utf8x]{inputenc}
\usepackage[T1]{fontenc}
\usepackage[english]{babel}
\usepackage{graphicx}
\usepackage[colorlinks=false]{hyperref}

\usepackage{hyphenat}
\usepackage[usenames,dvipsnames]{color}

\newcommand*{\atitle}[1]{\large \leavevmode{\textbf{#1}}}
\newcommand*{\authors}[1]{\normalsize \textbf{#1}}
\newcommand*{\affiliation}[1]{\footnotesize \leavevmode{#1}\\}
\newcommand*{\email}[1]{\leavevmode{\url{#1}}}

\captionnamefont{\footnotesize}
\captiontitlefont{\footnotesize}

\usepackage{enumerate}

\begin{document}
\vspace{6pt}
\begin{center}
\atitle{Four decades of circumpolar super-resolved satellite land surface temperature data}

\vspace{12pt}

\authors{Sonia Dupuis\textsuperscript{1,2}, Nando Metzger\textsuperscript{3}, Konrad Schindler\textsuperscript{3}, Frank Göttsche\textsuperscript{4}, Stefan Wunderle \textsuperscript{1,2}}

\vspace{6pt}

\affiliation{\textsuperscript{1}Oeschger Centre for Climate Change Research, University of Bern, Bern, Switzerland}
\affiliation{\textsuperscript{2}Institute of Geography, University of Bern, Bern, Switzerland}
\affiliation{\textsuperscript{3}Institute of Geodesy and Photogrammetry, ETH Zurich, Zurich, Switzerland}
\affiliation{\textsuperscript{4}Institute of Meteorology and Climatology Research, Karlsruhe Institute of Technology, Karlsruhe, Germany}

\end{center}
Corresponding author: Sonia Dupuis (sonia.dupuis@unibe.ch)
\vspace{16pt}

\begin{abstract}
Land surface temperature (LST) is an essential climate variable (ECV) crucial for understanding land-atmosphere energy exchange and monitoring climate change, especially in the rapidly warming Arctic. Long-term satellite-based LST records, such as those derived from the Advanced Very High Resolution Radiometer (AVHRR), are essential for detecting climate trends. However, the coarse spatial resolution of AVHRR’s global area coverage (GAC) data limit their utility for analyzing fine-scale permafrost dynamics and other surface processes in the Arctic. This paper presents a new 42 years pan-Arctic LST dataset,  downscaled from AVHRR GAC to 1 km with a super-resolution algorithm based on a deep anisotropic diffusion model. The model is trained on MODIS LST data, using coarsened inputs and native-resolution outputs, guided by high-resolution land cover, digital elevation, and vegetation height maps. The resulting dataset provides twice-daily, 1 km LST observations for the entire pan-Arctic region over four decades. This enhanced dataset enables improved modelling of permafrost, reconstruction of near-surface air temperature, and assessment of surface mass balance of the Greenland Ice Sheet. Additionally, it supports climate monitoring efforts in the pre-MODIS era and offers a framework adaptable to future satellite missions for thermal infrared observation and climate data record continuity.
\end{abstract}

\section*{Background \& Summary}

Land surface temperature (LST) is an important variable in the energy exchange between the land surface and the atmosphere. LST plays a key role in determining Earth's surface radiative energy budget and has been recognized as an Essential Climate Variable (ECV) by the Global Climate Observing System (GCOS)\cite{Bojinski2014} due to its relevance for climate change monitoring \cite{HULLEY201957,PEREZPLANELLS2023,Li2023}.
Satellite-based retrievals from thermal infrared (TIR) observations have enabled the characterization of LST at high spatiotemporal resolution, providing an indispensable metric for a wide range of environmental applications \cite{Li2023}. In the pan-Arctic region, LST is especially important for understanding the consequences of Arctic Amplification, which over recent decades has driven rapid warming  \cite{AMAPreport2021, Dada_2022, Rantanen2022}. Notably, permafrost temperatures across the northern high latitudes are responding to these warming trends \cite{Biskaborn2019}. LST can be used to model the thermal state of the ground and derive permafrost fractions \cite{ Westermann2017, obu_2019, Bartsch2023}, reconstruct two-meter air temperatures (T2M)  \cite{NielsenEnglyst2021} and provide input for surface mass balance models of the Greenland Ice Sheet \cite{Karagali2022}. Warming trends of LST are also associated with glacier ice retreat \cite{Gok2024}.

To detect statistically significant trends in ECVs, such as LST, a time series spanning at least 30 years is required \cite{WMO2010}. The considered period has to be sufficiently long to account for interannual variability \cite{IPCC_AR6_AnnexVII}. Long-term LST time series, of more than four decades of data (1980 to present), have been generated based on the heritage Advanced Very High Resolution Radiometer (AVHRR) mission \cite{essd-15-2189-2023, Ma2020}. 
A pan-Arctic AVHRR LST dataset based on EUMETSAT's global area coverage (GAC) fundamental data record (FDR) \cite{EUMETSAT2023} covering 1981-2020 has been specifically computed for the northern high latitudes \cite{tc-18-6027-2024}. To accurately capture small-scale variations in the ground thermal regime, studies suggest that a spatial resolution of 1 km or finer is necessary \cite{Bartsch2023,Nitze2018}: such LST data provide important information about ecosystems responses when exposed to extreme temperature \cite{Mildrexler2018}. In addition, changes about winter warming events in the Arctic would benefit from circumpolar high-resolution temperature datasets \cite{VikhamarSchuler2016}. The permafrost ECV products generated by the ESA Climate Change Initiative (CCI) Permafrost project are based on Moderate Resolution Imaging Spectroradiometer (MODIS) LST data, as they meet most user requirements, e.g. extensive geographical coverage, high spatial resolution (target resolution 1 km),  sufficiently high temporal resolution (monthly data) and multi-decadal temporal coverage (2000 - present) \cite{Bartsch2023CCI}. However, the MODIS timeseries  is less than the required 30 years. Therefore, a generated pan-Arctic AVHRR LST dataset with GAC resolution (called hereafter 'pan-Arctic AVHRR GAC LST'), extended to 2023, will be spatially enhanced to a spatial resolution of 1 km.

The most common approach for downscaling satellite-derived LST data is to perform spatiotemporal fusion (STF) on multiple sensors, such as Landsat or Advanced Spaceborne Thermal Emission and Reflection Radiometer (ASTER) and MODIS \cite{Wu2021,YOO2022102827}. STF generates a fine-spatial-resolution image by deriving conversion relationships from pairs of coarse-resolution and fine-resolution images, using either weighted function-based methods or learning-based methods \cite{Wu2021}. Other common methods include thermal sharpening and spectral unmixing-based algorithms \cite{agam_2007, Wu2012MODISTemporalFusion}. The main drawback of fusion  methods are inconsistencies that arise from differences in solar geometry, viewing angle and observation time and issues with non-homogeneous pixels for the thermal sharpening and unmixing methods \cite{Wu2021, liu_2007}.  Spatial downscaling is also a common task in the computer vision domain, where advanced deep-learning architectures have been exploited for image super-resolution (SR) or STF learning-based methods \cite{Sdraka2022}. Guided SR takes a low-resolution source image of some target quantity  as input, e.g., perspective depth acquired with a time-of-flight camera and a high-resolution guide image from a different domain, such as a greyscale image from a conventional camera\cite{delutio2022, Zhong2023}.  The target output is a high-resolution version of the source image. In Earth observation, guided super-resolution has been used to map above ground biomass or vegetation height with high-resolution satellite imagery as a guide  \cite{Karaman2025, delutio2019}. Previous work \cite{Dupuis2024} compared three strategies for downscaling AVHRR GAC LST data: a classical weight-based STF method, a learning-based STF method, and a guided super-resolution approach based on the deep anisotropic diffusion–adjustment (DADA) algorithm introduced by Metzger et al. (2023)\cite{Metzger2023}. The DADA algorithm is an edge-preserving filtering algorithm based on anisotropic diffusion \cite{perona} and guided by a separate guide image  \cite{daudt2019}. The DADA method  yielded the best results for downscaling AVHRR LST data. Building on these findings, the present study applies  DADA  to enhance the spatial resolution of the pan-Arctic AVHRR GAC LST dataset to 1 km.  The deep learning model is trained on MODIS LST data, where coarsened MODIS LST serve as  low-resolution input, and native-resolution MODIS LST provide the target. A three-channel guide image —composed of land cover data, a digital elevation model (DEM), and a vegetation height map— is used to inform the model. The main advantage of relying on static predictors is the ability to train a model at the circumpolar scale and to infer high-resolution LST across a long time-frame. During inference, the trained model downscales AVHRR GAC LST using the 1 km resolution guide, yielding a high-resolution pan-Arctic LST dataset suitable for permafrost modelling and Arctic climate analysis.

The objective is to deliver a twice-daily 1 km spatial resolution  LST time series for the entire pan-Arctic region covering more than four decades (1981-2023).   These data form a unique and highly valuable dataset for monitoring the high northern latitudes, which are particularly sensitive to climate change due to Arctic Amplification. The dataset provides valuable and so far missing information about LST dynamics in the pre-MODIS era (before 2000). Furthermore, alongside the LST product, the associated data used for the training of the super-resolution algorithm are also provided. The DADA super-resolution algorithm and its training data can be reemployed for similar tasks, which will be crucial for the upcoming thermal infrared missions to ensure sensor interoperability and the creation of long-term fundamental climate data records.
\section*{Methods}

\begin{figure}[ht]
\centering
\includegraphics[width=0.8\textwidth]{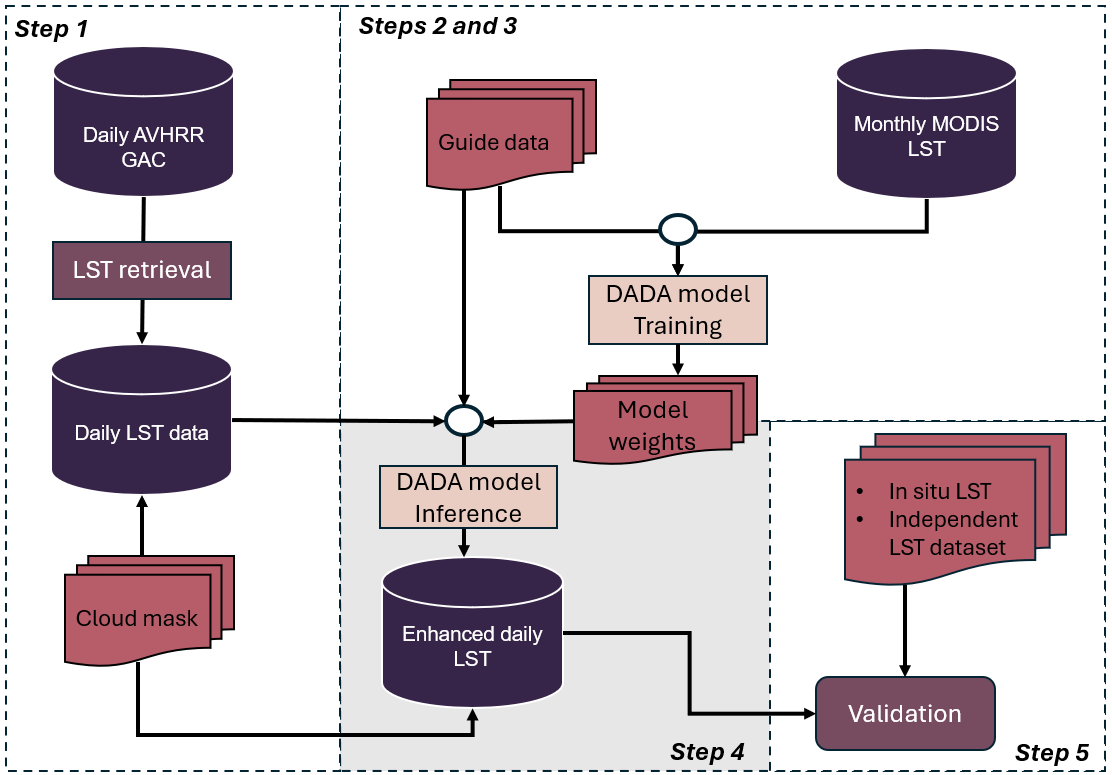}
\caption{ Downscaling workflow of the pan-Arctic AVHRR LST dataset.
}
\label{fig:workflow}
\end{figure}

The downscaling of the pan-Arctic AVHRR GAC LST dataset involves five main steps (Figure \ref{fig:workflow}) : (1) the computation of LST from AVHRR GAC brightness temperature \cite{tc-18-6027-2024}, (2) compilation of guide and MODIS datasets for the (3) training of the DADA deep learning model, (4) inference on AVHRR GAC data and (5) the validation and intercomparison of the final LST product. 

\subsection*{pan-Arctic AVHRR GAC LST dataset}

The pan-Arctic AVHRR GAC LST dataset is based on the EUMETSAT AVHRR GAC brightness temperature data for the period 1981-2023\cite{Dupuis2024}. The main difference with the previously generated LST GAC dataset is that the snow cover information has been updated to version 3.1 for the snow water equivalent (SWE) data \cite{Luojus2024} and to version 4.0 for the fractional snow cover (FSC) \cite{Luojus2022}.
The pan-Arctic AVHRR GAC LST dataset is a gridded product with extent (-180, 90, 180, 50\textdegree) distributed in the Network Common Data Form (NetCDF) format.
 Data from NOAA-7, 9, 11, 12, 14, 16, 17, 18 and 19, as well as the EUMETSAT MetOp satellites series (MetOp-A, MetOp-B and MetOp-C), are available. Twice daily data (day and night observations) are available for each satellite. NOAA-15 has been discarded due to data quality issues.

\subsection*{Datasets for DADA model training}
This study uses the multisensor LST IRCDR L3S ("MULTISENSOR\_IRCDR\_L3S\_0.01") and the AQUA Moderate Resolution Imaging Spectroradiometer (MODIS) LST L3C ("AQUA\_MODIS\_ L3C\_0.01") datasets developed by the ESA LST CCI project,  as well as auxiliary datasets such as a land cover from the ESA CCI on land cover, digital elevation model from Copernicus and a canopy height dataset developed at the ETHZ (details are given in the 'guide data' section). The ESA CCI LST datasets and the auxiliary dataset are used to train the deep learning neural network for the super-resolution task, as input source data and guide data respectively.
\subsubsection*{ESA CCI LST datasets}
The ESA CCI project on LST provides LST ECV products globally for the past 25 years \cite{Dodd2023, perez2023a}. Products based on different sensors are available, such as thermal infrared sensors and passive microwave sensors. All products follow strict validation procedures and are distributed as level 3 (L3) products with daily and monthly temporal frequency. The products are accessed through the python toolbox \textit{ ESA CCI Toolbox}, which provides directly access to the relevant data stores (\url{https://climate.esa.int/en/data/toolbox/}, last accessed: 15.09.2025).

The selected LST datasets for this study are based on Aqua-MODIS with observation time at approximately 1:30 pm and 1:30 am (local solar time) and the multi-sensor climate data record (CDR) constructed from the ATSR-2, AATSR, Terra-MODIS and Sentinel-3 Sea and Land Surface Temperature Radiometer (SLSTR) instruments. These instrument series all have observation times at roughly 10:30 am and 1:30 pm (local solar time). The LST retrieval is performed with the University of Leicester (UOL) algorithm for the IRCDR dataset and the generalized split-window (GSW) algorithm \cite{Wan1996} for the Aqua-MODIS dataset \cite{Dodd2023}. The UOL algorithm introduces a dependence on land cover: land surface emissivity (LSE) is implied in the retrieval coefficients \cite{Prata2002}. The GSW algorithm, on the other hand, assumes that LSE is known a priori.

\subsubsection*{Guide data}
The super-resolution of the pan-Arctic AVHRR GAC LST dataset requires auxiliary datasets. The following datasets are included in the guide:
\begin{itemize}
    
\item The Copernicus digital elevation model (DEM) GLO-90, upscaled to 0.01° spatial resolution (\url{https://doi.org/10.5270/ESA-c5d3d65}\nocite{dem}, last accessed: 15.09.2025). This dataset represents the elevation of the Earth's surface and is derived from interferometric synthetic aperture radar (InSAR) data acquired during the TanDEM-X Mission \cite{Zink_tandem}.

\item ESA CCI Land Cover for the year 2005: The land cover data comes from the ESA  CCI Land Cover project, which provides global land cover maps derived from MERIS global surface reflectance and SPOT vegetation datasets. Each pixel in the land cover map corresponds to a land cover class labelled according to the UN Land Cover Classification System (LCCS), which includes 22 global land cover categories. The LCCS classifiers also enable the conversion to Plant Functional Types (PFTs), which are used in Earth System Models (\url{https://www.esa-landcover-cci.org}, last access: 28 July 2025). The spatial resolution of the original data was upscaled to 0.01° to match the MODIS LST dataset. This upscaling was achieved by selecting the most frequent land cover class within each upscaled pixel, ensuring consistency with the original land cover distribution.

\item A high-resolution canopy height model derived from the global ecosystem dynamics investigation (GEDI) space-borne LiDAR mission and satellite data from Sentinel-2 \cite{Lang2023}. The spatial resolution was decreased to match the spatial resolution of the MODIS LST datasets (0.01°).
\end{itemize}

\subsubsection*{Data sampling}

\begin{figure}[ht]
\centering
\includegraphics[width=\linewidth]{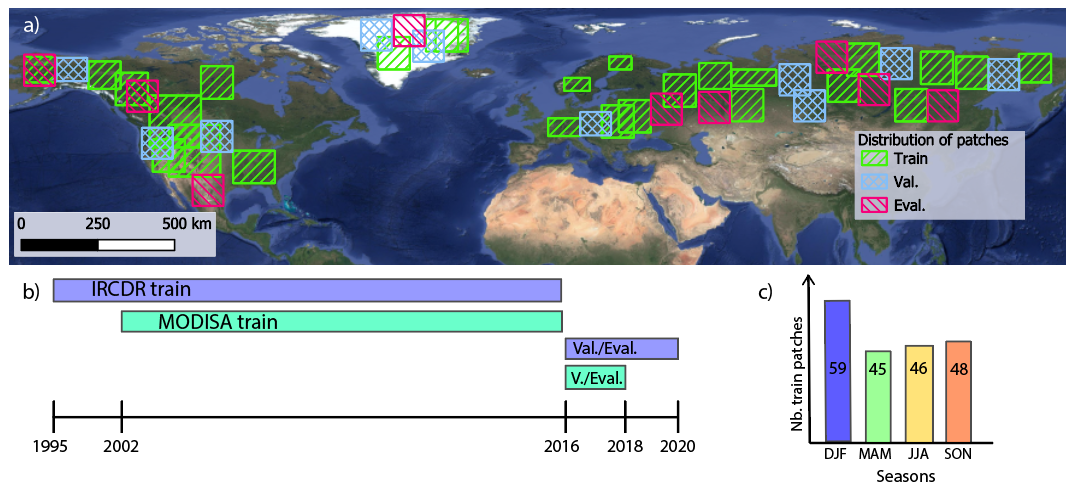}
\caption{ Overview of selected scenes for the training of the super-resolution algorithm. The spatial distribution of training, validation and evaluation scenes are shown in a), the temporal span of all sets for both LST datasets is shown in b) and the count per season for the scenes belonging to the training set are shown in c).
}
\label{fig:trainingscenes}
\end{figure}

To avoid the persistent cloud cover in the Arctic, monthly mean LST scenes with as little cloud cover as possible are selected from the ESA CCI LST datasets across the pan-Arctic region, the Alps and the United States of America (USA). For the model training, 198 scenes with a size ranging from 1600$\times$1600 pixels to 400$\times$700 pixels are selected. The size varies to avoid as many  missing pixels as possible due to inland water bodies and coastal zones, as water bodies are masked out in the LST CCI datasets. These different scenes have been selected to cover diverse topographic features, diverse types of land cover and a diverse temperature range, which is obtained by sampling different  Köppen–Geiger climate classes \cite{Beck2018} separately. Additionally, the scenes were sampled to cover day and night times as well as different seasons and different years. The training data covers the period from 1995 to 2015 for the IRCDR and 2002 to 2015 for the MODISA dataset, and the validation and evaluation set covers the years 2016 to 2018, or 2020 for the IRCDR. The validation set contains 20 scenes, and the evaluation set contains eight scenes. The validation scenes have a size of 960$\times$960 pixels, and the evaluation scenes all have a size of 1000$\times$1000 pixels. An overview of the data sampling is presented in Figure \ref{fig:trainingscenes}. Additionally, the samples scenes as well as the auxiliary data are available from Zenodo: \url{https://zenodo.org/records/17341544}.

\subsection*{DADA model training}
\begin{figure}[ht]
\centering
\includegraphics[width=0.8\textwidth]{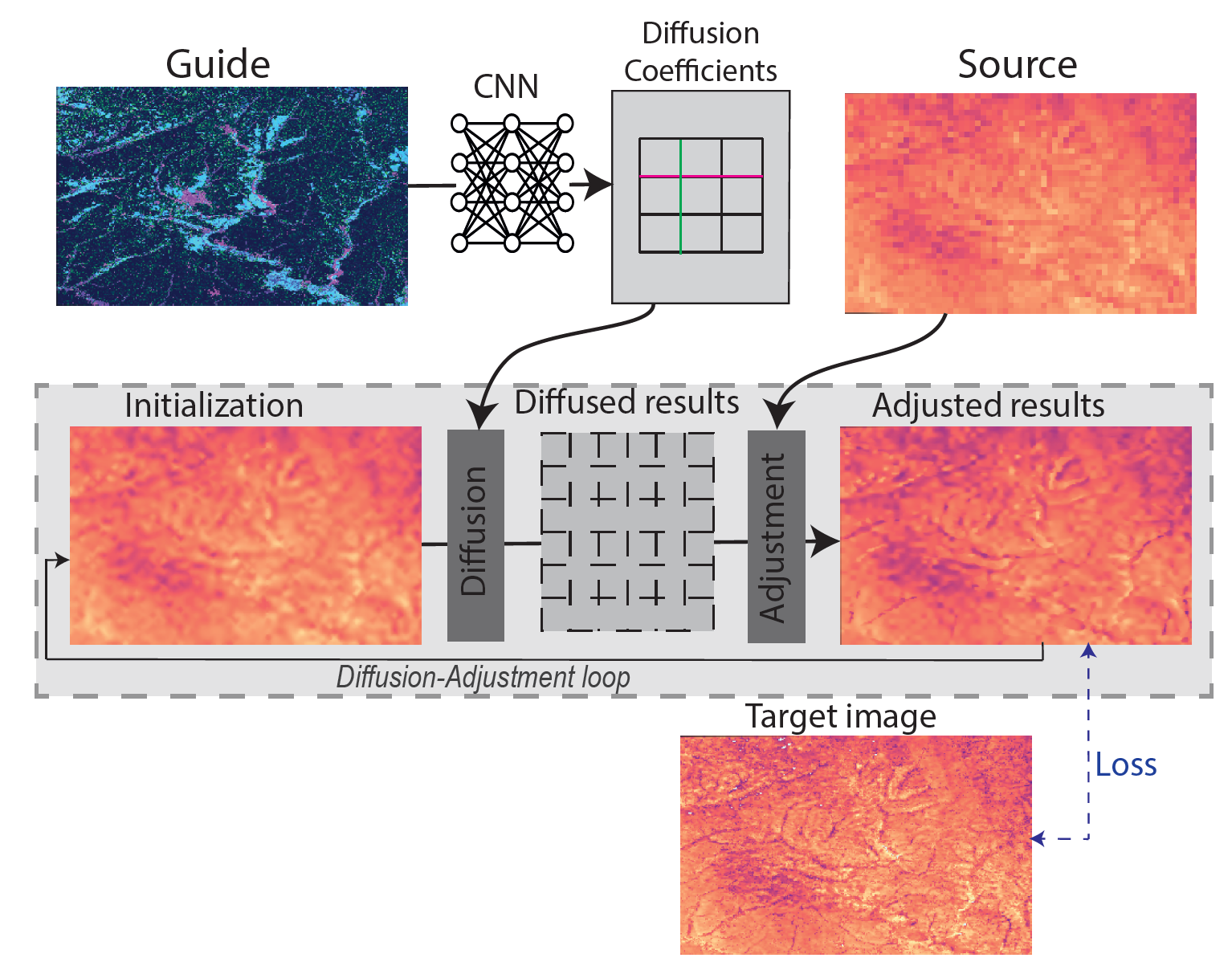}
\caption{ Overview of the algorithm. The source is coarsened MODIS data, and the target image is the original MODIS monthly mean LST information. Adapted from Metzger et al., (2023) \cite{Metzger2023}.
}
\label{fig:dada_schema}
\end{figure}

The framework for guided super-resolution as presented in Metzger et al. (2023) \cite{Metzger2023}, written in the PyTorch framework\cite{PyTorch}, has been adapted to integrate the TorchGeo python library \cite{Stewart_TorchGeo_Deep_Learning_2024} to handle geospatial data. An overview of the DADA model training is shown in Figure \ref{fig:dada_schema}. The neural feature extractor is, as in the original work, a U-Net with ResNet-50 backbone \cite{Metzger2023}. ResNets with different depths (18 and 34) were tried but ResNet-50 presented better performances on the evaluation dataset (see Table \ref{table:tab1}). 

All sampled scenes contained in the training, validation and evaluation sets were extracted from the NetCDF data cubes, available from the \textit{ ESA CCI Toolbox} and saved as separate GeoTIFF files to ensure seamless integration with TorchGeo. Before the actual training of the model, the guide dataset is created by taking the intersection of the land cover, DEM and vegetation height datasets. Then, the newly created three-channel guide is intersected with the LST scenes. This ensures that for each LST scene, a corresponding guide scene is assigned and that only LST scenes with valid guide data are selected.

During the training phase, data augmentation in the form of random horizontal and vertical flipping is applied. Both LST and guide data are previously scaled with a min-max scaler. In addition, random Planckian color jittering is applied to the guide \cite{zini2022planckian}. During training, each epoch consists of 400 randomly sampled patches of size 240 $\times$ 240 pixels, drawn from the training dataset. With a batch size of 8, this corresponds to 50 iterations per epoch. The model converges after approximately 3000 such epochs (i.e., 150,000 total iterations, see Figure \ref{fig:iterations}). The original hyperparameters from Metzger et al., (2023) have been kept, with exception of the learning rate scheduler step size  (\texttt{lr\_step}), which is set to 150. Performances for a selection of alternative settings are shown in Table \ref{table:tab1}.  The ESA CCI LST data is coarsened $\times$5 during training with NaN-aware average pooling, and these coarsened patches represent the source image and the original ESA CCI LST patch is the target image. The scaling factor is $\times$5 to downscale the AVHRR scenes, with 0.05 \textdegree{} spatial resolution, to the 0.01 \textdegree{} resolution of MODIS LST. Lakes, rivers and smaller inland water bodies are masked out in the ESA CCI LST datasets, therefore a few holes are present in the patches. The algorithm is able to handle gaps due to clouds and missing pixels \cite{Metzger2023}.

During the validation phase, the same scaling is applied to the datasets. This time, the scenes from the validation dataset are sampled systematically by extracting patches of 240 $\times$ 240 pixels in a grid-like fashion for the entire area of interest. Exactly 16 patches of 240 $\times$ 240 pixels are fed into the model for each image. During the model evaluation, only inference is performed, and patches are again selected systematically with a smaller stride parameter. The smaller stride parameter allows an overlap in the output evaluation patches that are merged for the analysis. This overlap avoids stitching issues in the final product \cite{Huang2018}. The final product is obtained by computing the mean LST value for all pixels presenting overlapping patches.

\subsection*{DADA model evaluation}

The performance of the model was evaluated using independent geographical patches (Figure \ref{fig:trainingscenes}) that capture both daytime and nighttime temperature dynamics across the pan-Arctic region. The spatial distribution of residuals and their corresponding error histograms are shown in Figures \ref{fig:Figure_residuals} and \ref{fig:Hist_residuals}. The error distribution is  centred around 0 °C for all regions with a standard deviation ranging from 0.914°C to 1.331°C. Residuals exceeding a few degrees can be attributed to boundary effects at sharp transitions, such as the edges of water bodies or small gaps caused by cloud coverage. These effects are particularly visible near the bottom in the Krasnoyarsk panel in Figure \ref{fig:Figure_residuals}. Larger residuals (> 5°C) occur very rarely and are mainly associated with artefacts in the satellite imagery. In general, abrupt temperature gradients that are not well represented by the guide features tend to produce higher errors. For instance, due to its spatial homogeneity in the guide features the Greenland Ice Sheet (Greenland panel, Figure \ref{fig:Figure_residuals}) exhibits larger discrepancies, which cannot reproduce fine-scale temperature variations in the target data.

\begin{figure}[ht]
\centering
\includegraphics[width=0.9\textwidth]{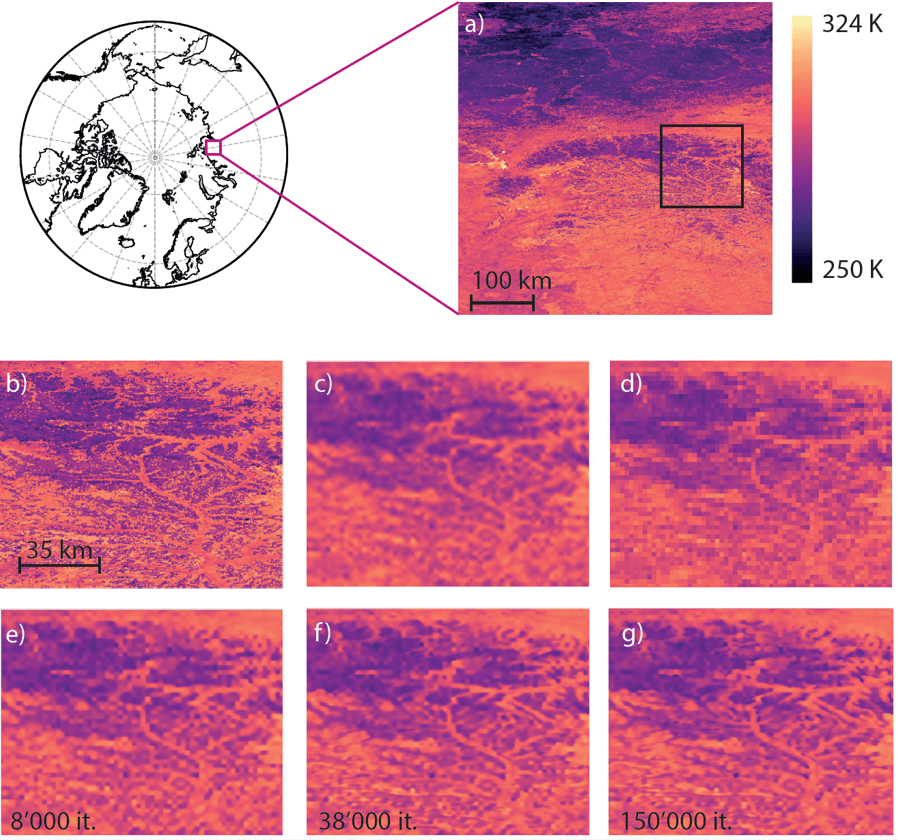}
\caption{ Evaluation example for an evaluation scene in Northern Siberia in June 2018 (local equatorial crossing time = 10:30 am). a) Shows the entire evaluation scene (original MODIS LST) with a spatial subset indicated by a black box, that is used in subsets b) to g). b) presents the original MODIS LST scene (the target). c) represents the bicubic interpolation and d) the coarsened MODIS scene (the source). e), f) and g) represent the evaluation results at 8000, 38000 and 150000 iterations, respectively.
}
\label{fig:iterations}
\end{figure}

\begin{figure}[ht]
\centering
\includegraphics[width=\linewidth]{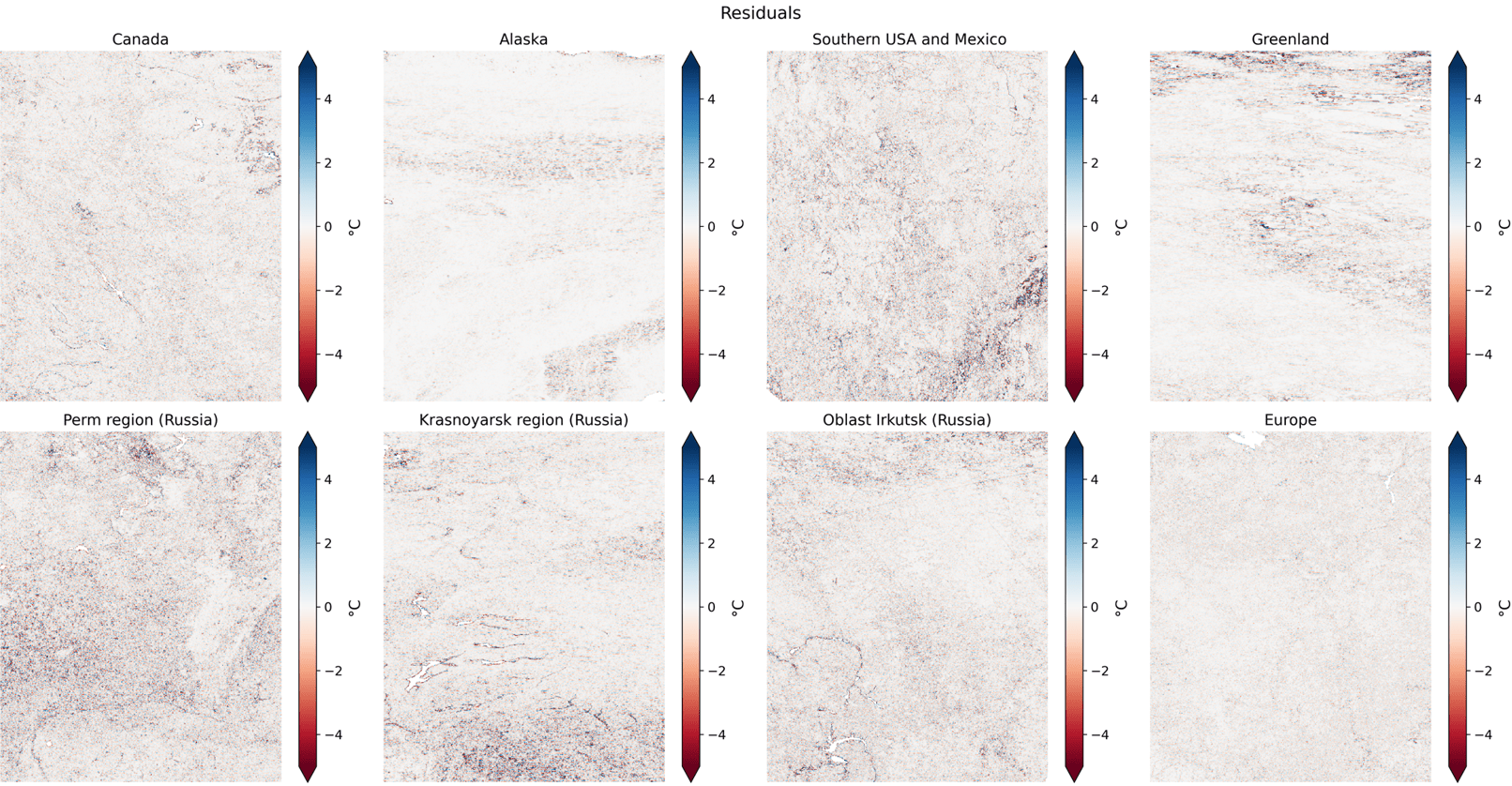}
\caption{Difference maps for the eight evaluation scenes (original MODIS LST - inferred MODIS LST). Blue denotes underestimation and red overestimation of the LST.
}
\label{fig:Figure_residuals}
\end{figure}

\begin{figure}[ht]
\centering
\includegraphics[width=\linewidth]{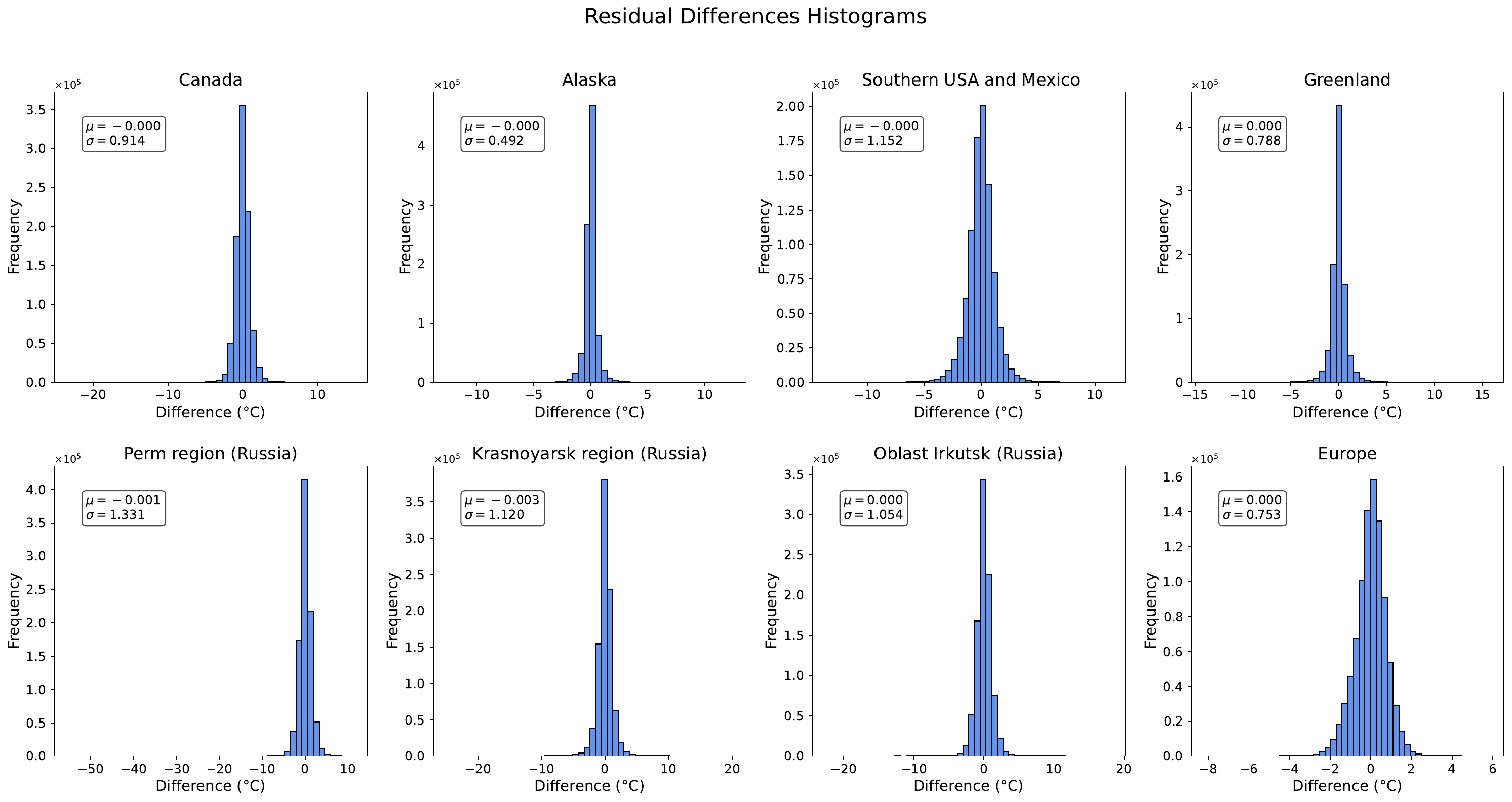}
\caption{Histograms of residuals (original MODIS LST - inferred MODIS LST) for geographical patches used for model evalutation (see Figure \ref{fig:Figure_residuals}).
}
\label{fig:Hist_residuals}
\end{figure}

The overall model performance, quantified by the mean absolute error (MAE) and root mean square error (RMSE), is summarized in Table \ref{table:tab1}. The values show the overall results, averaging all eight evaluation scenes. The final configuration employs both ESA LST CCI datasets for training, a ResNet-50 backbone, a sampler length of 400, and \texttt{lr\_step} of 150. To further assess the influence of different training datasets (MODISA and IRCDR), additional experiments were conducted using each dataset independently, as well as with daytime-only and nighttime-only subsets. Figure \ref{fig:iterations} illustrates the evaluation results for a monthly mean daytime Terra-MODIS LST scene in Northern Siberia, including the target image, the source input, bicubic interpolation results, and model outputs across several training iterations until convergence.

The MAE and RMSE values reported in Table \ref{table:tab1} are very similar, differing only by fractions of a degree. This consistency, over an extensive evaluation dataset (8 million pixels) that covers many local temperature patterns and variations, demonstrates the robustness of the model to architectural and training changes. STF models often do not generalize well across regions due to uncertainties of spatial, temporal, and spectral patterns inherent to thermal imaging \cite{HUANG2024102505}. However, the spatial residual patterns (Figure \ref{fig:Figure_residuals}) show no major discrepancies between regions with contrasting topography and climate. The final MAE and RMSE values are consistent with the accuracy levels of the pan-Arctic AVHRR LST GAC and ESA LST CCI datasets when compared to in-situ observations \cite{Dupuis2024, LSTCCI_PVIR_2023}.

\begin{table}
\centering
\caption{Mean MAE and MSE of the eight evaluation scenes (Figure 5) for selected runs. Unless otherwise stated, results are obtained with ResNet-50, \texttt{lr\_step=150} and \texttt{sampler\_length=400}.}
\begin{tabular}{lcc}
\hline
\textbf{Run} & \textbf{Mean MAE [$^\circ$C]} & \textbf{Mean RMSE [$^\circ$C]} \\
\hline
\multicolumn{3}{l}{\textbf{Default (ResNet-50, lr\_step=150)}} \\
All data                 & 1.150 & 2.297 \\
IRCDR data only          & 1.159 & 2.313 \\
MODISA data only         & 1.173 & 2.336 \\
NIGHT data only          & 1.174 & 2.336 \\
DAY data only            & 1.168 & 2.327 \\
Bicubic                  & 1.244 & 2.396 \\
Source (coarse MODIS)    & 1.283 & 2.440 \\
\hline
\multicolumn{3}{l}{\textbf{ResNet Variants}} \\
ResNet-18, lr\_step=150  & 1.160 & 2.313 \\
ResNet-34, lr\_step=150  & 1.158 & 2.310 \\
ResNet-50, lr\_step=100  & 1.158 & 2.312 \\
ResNet-50, lr\_step=300  & 1.156 & 2.319 \\
sampler\_length = 800    & 1.157 & 2.315 \\
\hline
\end{tabular}
\label{table:tab1}
\end{table}

\clearpage
\subsection*{Inference on AVHRR data}

The pan-Arctic AVHRR GAC LST dataset with 4 km spatial resolution (GAC format) is stored in  NetCDF-4 format. Therefore, the inference workflow for the AVHRR data has been built with the Python package Xarray \cite{Hoyer2017, TorchGeoGeoPy},  to read in labelled multi-dimensional arrays in the NetCDF format. Before inference, cloud masking is applied to the AVHRR GAC LST scenes. The downscaled LST scenes are then computed as patches of 1920$\times$1920 pixels with a stride of 1480 vertically and 1792 horizontally. For each AVHRR scene covering the pan-Arctic, defined as the part of the globe north of 50\textdegree, 40 patches are computed. The patches are stitched together for each timestamp by taking the average value in the presence of overlapping pixels, the same cloud mask is applied as for the corresponding LST GAC images, and performance and quality indicators from the LST GAC images are applied to the downscaled images. The final data are stored in  NetCDF-4 format with accompanying metadata. The filename and the metadata of each NetCDF file contain the timestamp, the satellite name and the processing version.

\section*{Data Record }

The downscaled version of the pan-Arctic AVHRR GAC LST dataset\cite{dupuis2025} (hereafter 'AVHRR SR LST') is accessible from the data portal of the University of Bern (Boris Portal): \url{https://boris-portal.unibe.ch/entities/product/761f8e2f-fb77-4efc-beaf-d196c000ffea}. The original AVHRR GAC dataset and the downscaled version are distributed in the same dataset, under different variables. The entire dataset is called hereafter 'pan-Arctic AVHRR LST dataset'.
The ESA CCI LST scenes and auxiliary data \cite{dupuis_2025_modis} used to train the DADA model are available from Zenodo: \url{https://doi.org/10.5281/zenodo.17341544}.
The AVHRR SR LST  are stored in NetCDF-4 format files and  provide twice daily clear-sky LST (at day and night time) for each satellite, covering a total of 42 years (1982 to 2023) at 0.01° spatial resolution. Clouds are masked out from the data and are replaced with NaN values.  Water bodies have not been masked out, due to the high classification uncertainty of water bodies in the high Arctic \cite{Bartsch2024CircumarcticGradients}, and to let users choose the right water mask for their purpose. The data are provided as a gridded lat-lon product, in  WGS84 projection (EPSG:4326) with extent -180°- 180° longitude and 50°-90° latitude. Files containing large gaps due to scanline failures or other data outages have been filtered out. The dataset is organized by year and  satellite, each folder containing 730 NetCDF files for a complete year (732 for leap years). An overview of the available data is provided in Table \ref{tab:AVHRR  coverage}.

 Filenames are encoded with the timestamp, satellite name, version name and a DAY or NIGHT tag. The files can easily be opened in Python with the xarray package, either separately or directly in a data cube, by reading a whole folder simultaneously. Each NetCDF file contains distinct data arrays for LST, acquisition time, satellite zenith angle, performances of the GSW algorithm as well as sun zenith angle. The  key variables, stored as data arrays in NetCDF-4 format, are listed in the Table  \ref{tab:variables}. All data arrays have associated metadata. The AVHRR SR LST data are stored under the variable 'LST' and the original LST data at GAC resolution are stored under the variable 'LST-GAC'. Cloudy pixels in the GAC dataset are assigned the value - 110. Satellite overpass time is saved in variable 'scanline time' as fractional hours of the day and satellite and sun zenith angle information are also available. 'Test MAE' and R\textsuperscript{2} values reflect the  performances of the GSW algorithm and can be used as proxies for data quality.
 
 The files are stored in a compressed format using a scale factor and an offset value. Most GIS tools and geodata packages automatically apply these formatting values at data opening. However, they have to be applied  manually when creating custom parsing routines.

 \begin{table}
\centering
    \caption{Data coverage of the pan-Arctic AVHRR LST dataset per satellite}

    \begin{tabular}{ccccc}
        \hline
        \textbf{Instrument} & \textbf{Satellite} & \textbf{Data start} & \textbf{Data end} & \textbf{Node: Daytime overpass time}\\
        \hline
        AVHRR/2 & NOAA-7& 1981-08-24 & 1985-01-31 & Ascending: PM \\
        AVHRR/2 & NOAA-9& 1985-02-25 & 1988-11-07 & Ascending: PM\\
        AVHRR/2 & NOAA-11& 1988-11-08 & 1994-09-13 & Ascending: PM\\
        AVHRR/2 & NOAA-12& 1991-09-16 & 1998-11-26 & Descending: PM\\
        AVHRR/2 & NOAA-14& 1995-01-20 & 2002-08-01 & Ascending: PM\\
        AVHRR/3 & NOAA-16& 2001-01-01 & 2009-12-31 & Ascending: PM\\
        AVHRR/3 & NOAA-17& 2002-07-10 &  2010-02-28 & Descending: AM\\
        AVHRR/3 & NOAA-18& 2005-06-05 & 2016-12-31 & Ascending: PM\\
        AVHRR/3 & NOAA-19& 2009-02-22 & 2019-12-31 & Ascending: PM\\
        AVHRR/3 & MetOp-A& 2007-06-29 & 2021-11-27 & Descending: AM\\
        AVHRR/3 & MetOp-B& 2013-01-01 & 2023-12-31& Descending: AM \\
        AVHRR/3 & MetOp-C& 2019-07-01 & 2020-12-31& Descending: AM\\
      \hline
    \end{tabular}
    
        \label{tab:AVHRR  coverage}
\end{table}

 \begin{table}
    \caption{NetCDF data variables in the pan-Arctic AVHRR LST product files.}
    \resizebox{\textwidth}{!}{%
    \begin{tabular}{p{2.1cm}p{4cm}ccccc}
        \hline
        \textbf{Variable name} & \textbf{Long name} & \textbf{Standard name} & \textbf{Units} & \textbf{Valid range} & \textbf{Scale factor} & \textbf{Offset} \\
        \hline
        LST & enhanced daytime land surface temperature (0.01° GSD)  & lst & K & 200-360 & 0.01 & 273.15\\
        LST\_GAC & daytime land surface temperature (0.05° GSD) & lst\_gac & K & 200-360  & 0.01 & 273.15\\
        scanline\_time & 
scanline time as fractional hours of the day & time & h & 0-24000  & 0.01 & 0.0\\
        satzen & Satellite Zenith Angle & sensor\_zenith\_angle &degrees & 0-18000 & 0.01&0.0\\
        sunzen & Sun Zenith Angle & solar\_zenith\_angle &degrees & 0-7500 &0.01 & 0.0\\
        test\_mae & Mean Absolute Error (MAE) value of GSW algorithm performances - evaluation on test set & test\_mae & K & - & 0.01 & 0.0 \\
        r2 & R\textsuperscript{2} value of GSW algorithm performances - evaluation on test set &  R\textsuperscript{2}&  - & - & 0.01& 0.0\\

      \hline
    \end{tabular}
    } 

    \label{tab:variables}
\end{table}

\clearpage

\section*{Technical Validation }

\subsection*{Validation against in situ measurements}
The most conventional and reliable method to validate satellite-derived LST is the 'T-based' validation, which directly compares in situ LST at the satellite overpass \cite{Li2023, Gottsche2016, perez2023b}.
 In situ LST obtained from the Surface Radiation Budget (SURFRAD) network, the Karlsruhe Institute of Technology (KIT) network  \cite{Gottsche2016, Martin2019}, the Atmospheric Radiation Measurement Climate Research Facility US Department of Energy (ARM) site at the North Slope of Alaska (NSA), the “Copernicus Space Component Validation for Land Surface Temperature, Aerosol Optical Depth and Water Vapor Sentinel-3 Products Project“ (LAW project, \url{https://law.acri-st.fr/home}, last accessed: 20-09.2025) and the Baseline Surface Radiation Network (BSRN) \cite{maturilli2020baom, cox2021bmor, kustov2018baom, kustov2023baom}  are used for validation. Table \ref{tab:insitu_stations} lists the stations: the KIT stations, which are part of LSA SAF’s validation effort and supported by EUMETSAT, are located in different climate zones, such as Mediterranean, marine west coast and humid continental climates \cite{Gottsche2016}.\nocite{surfrad,kit_network, arm} In situ LST is computed from measured radiation components as in Martin et al., (2019)\cite{Martin2019}. BSRN stations presenting a sufficiently long data record for validation have been chosen to complement the coverage of in situ stations in the Arctic.

\begin{figure}
\centering
\includegraphics[width=\linewidth]{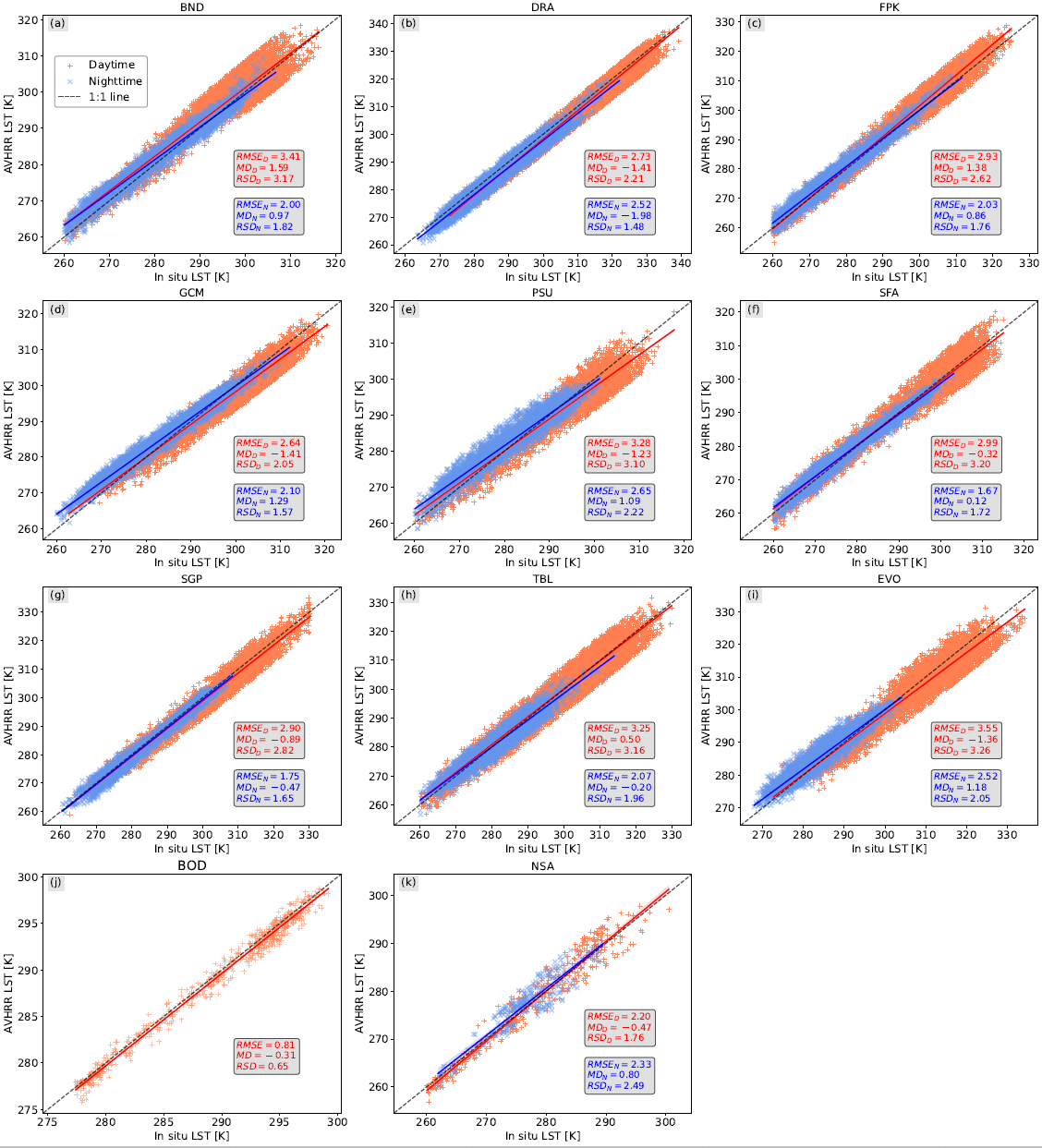}
\caption{Enhanced (1 km) AVHRR LST versus in situ LST at (a) Bondville (BND), (b) Desert Rock (DRA), (c) Fort Peck (FPK), (d) Goodwin Creek (GCM), (e) Penn. State Univ (PSU), (f) Sioux Falls (SFA), (g) Southern Great Plains (SGP), (h) Table Mountain (TBL), (i) Evora (EVO), (j) Lake Constance (BOD) and (k) North Slope of Alaska (NSA). Red represents daytime and blue nighttime measurements. Match-up periods differ and are provided in the text.
}
\label{fig:Figure_03_SURFRAD_insitu}
\end{figure}

\begin{figure}
\centering
\includegraphics[width=\linewidth]{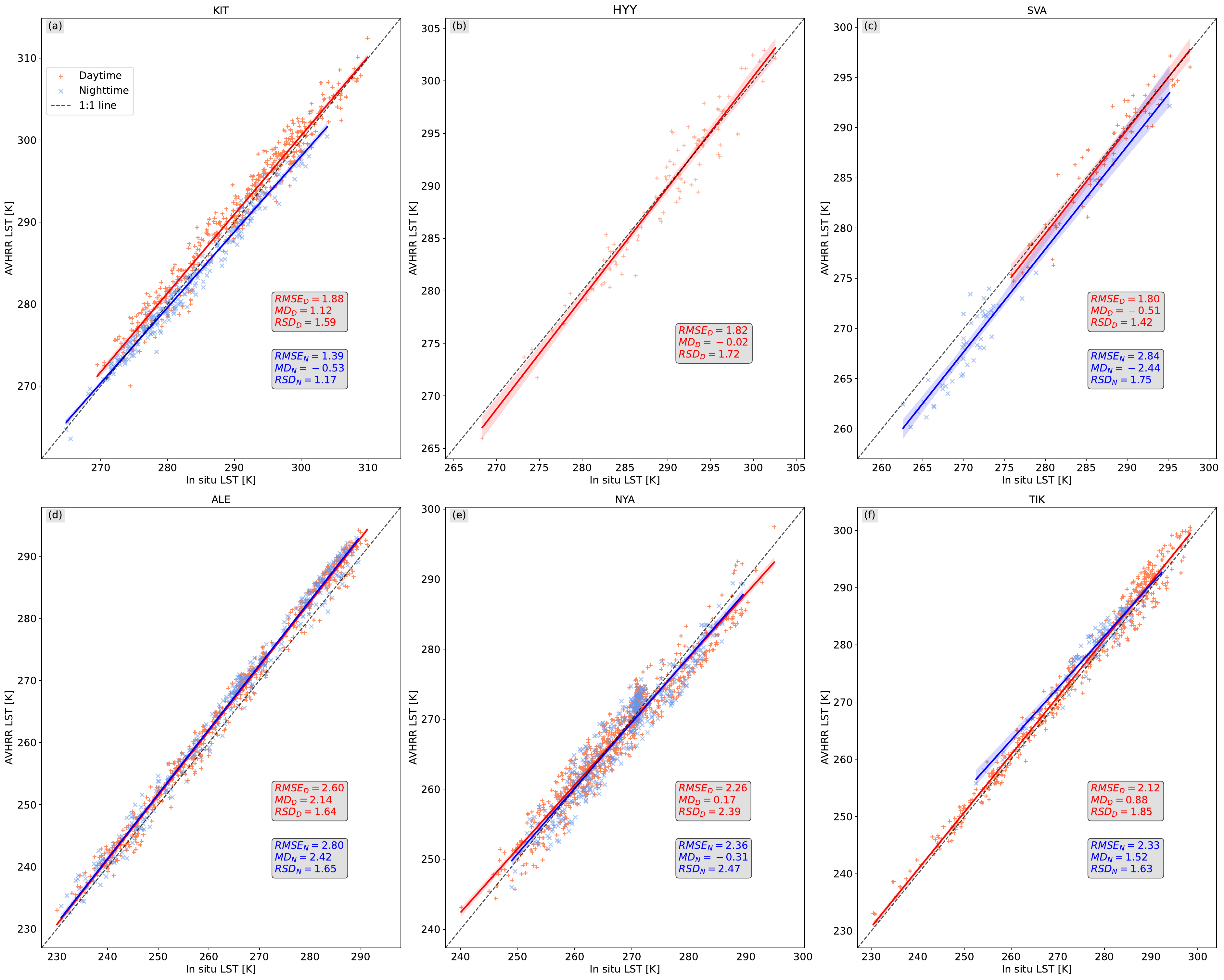}
\caption{Enhanced (1 km) AVHRR LST versus in situ LST at (a) the KIT Campus Nord (Germany), (b) Hyytiälä,  (c) Svartberget (Sweden), (d) Alert (Canada), (e) Ny-Ålesund (Norway) and (f) Tiksi (Russia) . Red represents daytime and blue nighttime measurements. Match-up periods differ and are provided in the text.
}
\label{fig:Figure_04_LAW_results}
\end{figure}

 Figures \ref{fig:Figure_03_SURFRAD_insitu} and \ref{fig:Figure_04_LAW_results} show the validation results for NOAA-14, 16, 17, 18 and 19, and MetOp-A, B and C against in situ LST from the
validation sites in Table \ref{tab:insitu_stations}. The validation is separated for daytime (represented in red) and nighttime (blue) observations. For each validation
site, the median deviation (MD), RMSE and robust standard deviation (RSD), as computed in Pérez-Planells et al. (2023)\cite{perez2023a} are shown. The match-up with the SURFRAD stations covers the period
from 1985 to 2023, the match-up period for KIT station in EVORA (EVO) starts in 2009 and ends in 2023,  the match-up period for Lake Constance starts in 2016 and ends in 2020, and  the match-up period for the ARM site  North Slope of Alaska (NSA) covers 2007-2014 and 2020-2023. Regarding the BSRN data, the match-up period  for  station Alert (ALE) starts in 2004 and ends in 2014, Ny-Ålesund (NYA)  starts in 2006 and ends in 2020 and Tiksi (TIK) starts in 2010 and ends in 2018. The LAW stations (KIT, HYY and SVA) cover the period 2021 to 2023. The obtained results are very similar to the results obtained with the GAC product \cite{tc-18-6027-2024}, the present results are very similar. This can be attributed to the fact that the chosen  SURFRAD stations are located in relatively homogeneous areas, i.e. regions where land surface properties such as vegetation, topography, and land cover vary little across space. As a result, the difference in spatial resolution between the 4 km and 1 km products has  limited impact.
Compared to previous studies on AVHRR LST \cite{essd-15-2189-2023, Ma2020, Reiners2021}, the present dataset shows similar accuracy and precision.

\begin{table}
    \caption{Measurement stations used for LST validation. Station name and ID,  network the station belongs to, latitude, longitude, elevation above sea level and the dominant land cover type.}
   \resizebox{\textwidth}{!}{
    \begin{tabular}{p{4.5cm}ccccp{3cm}}
        \hline
        \textbf{Station name (ID)} & \textbf{Network} & \textbf{Latitude [°]} & \textbf{Longitude [°]} & \textbf{Elevation [m]} & \textbf{LCCS} \\
        \hline
        Bondville, Illinois (BND) & SURFRAD & 40.0519 & -88.3731 & 230 & Cropland \\
        Desert Rock, Nevada (DRA) & SURFRAD & 36.6237 & -116.0195 & 1007 & Open Shrubland \\
        Fort Peck, Montana (FPK) & SURFRAD & 48.3078 & -105.1017 & 634 & Grassland \\
        Goodwin Creek, Mississippi (GCM) & SURFRAD & 34.2547 & -89.8729 & 98 & Wooded Grassland \\
        Penn. State Univ., Pennsylvania (PSU) & SURFRAD & 40.7201 & -77.9309 & 376 & Deciduous Broadleaf Forest \\
        Sioux Falls, South Dakota (SFA) & SURFRAD & 43.73403 & -96.62328 & 1689 & Cropland \\
        ARM Southern Great Plains, Oklahoma (SGP) & SURFRAD & 36.60406 & -97.48525 & 314 & Cropland \\
        Table Mountain, Boulder, Colorado (TBL) & SURFRAD & 40.1250 & -105.2368 & 1689 & Cropland \\
        Lake Constance, Germany (BOD) & KIT & 47.58 & 9.57 & 396 & Water \\
        Evora, Portugal (EVO) & KIT & 38.54 & -8.003 & 300 & Mosaic Tree and Shrubs \\
        North Slope of Alaska, USA (NSA) & ARM & 71.323 & -156.609 & 8 & Lichens and Mosses \\
        Alert, Canada (ALE) & BSRN & 82.49 & -62.42 & 127 & Bare Soil \\
        Ny-Ålesund, Norway (NYA) & BSRN & 78.9227 & 11.9273 & 11 & Bare soil \\
        Tiksi, Russia (TIK) & BSRN & 71.5862 & 128.9188 & 48  & Shrubland \\
         Svartberget, Sweden (SVA) & LAW & 64.26 & 19.77 & 269 & Mixed Forest \\
        Hyytiälä, Finland (HYY) & LAW & 61.85 & 24.29 & 181 &  Mixed Forest \\
        KIT Forest, Germany (KIT) & LAW & 49.09 & 8.43 & 115  & Deciduous Broadleaf Forest \\
      \hline
    \end{tabular}
    } 

    \label{tab:insitu_stations}
\end{table}

\clearpage

\subsection*{Intercomparison with EDLST from LSA SAF}
Due to differences in satellite equatorial crossing times, which would lead to large differences in LST or diverging compositing strategies, such as mean daily composites versus selection of the scene nearest to nadir, only qualitative comparisons against other LST satellite products with a spatial resolution of 0.01° are possible.

The satellite application facility on land surface analysis (LSA SAF), as part of the EUMETSAT application ground segment, delivers products and services for land surface monitoring, such as albedo, LST and emissivity, surface radiation, vegetation and wildfires. The LSA SAF generates and distributes a LST product derived from AVHRR onboard the EUMETSAT polar system satellites, the MetOp series. This  EUMETSAT Polar System (EPS) daily LST prodcut (EDLST; LSA-002), is available from 2015 until the present and disseminated in a global grid in sinusoidal projection with a spatial resolution of 0.01° \cite{Trigo2017}. The EDLST product is generated with the GSW algorithm \cite{Wan1996}, which is calibrated with data from the SeeBor database \cite{Borbas2005} and the radiative transfer simulations using the MODerate spectral resolution atmospheric TRANSsmittance algorithm (MODTRAN4) software \cite{Berk2000}.

\begin{figure}
\centering
\includegraphics[width=\linewidth]{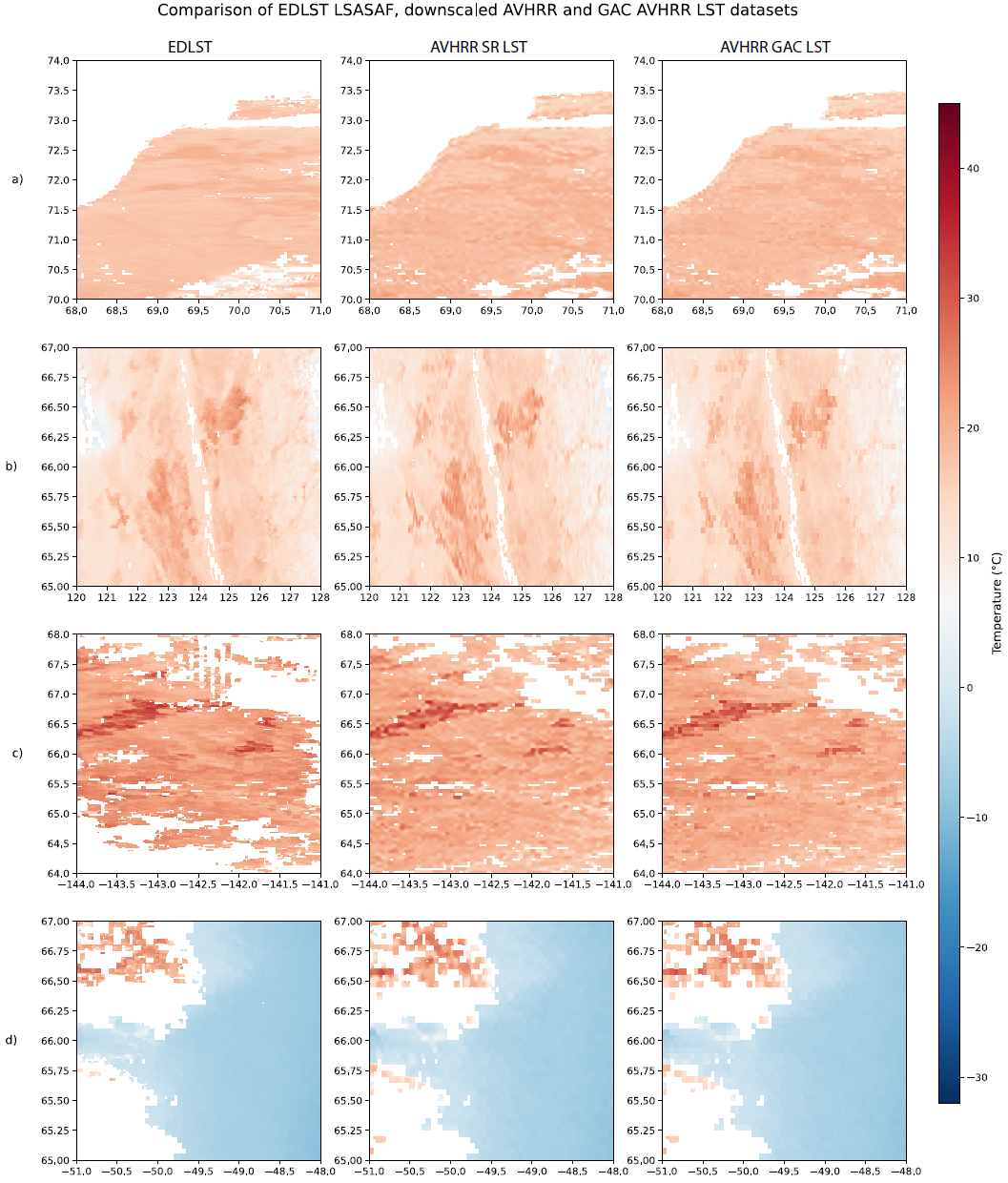}
\caption{Comparison of the  EDLST product and the AVHRR SR LST dataset from the MetOp-A satellite during daytime. a) On the 20.08.2020, in Northern Siberia, b) on the 20.08.2020 in East Siberia, c) on the 10.06.2020 in Alaska and d) on the 20.05.2020 in Greenland. The cloud mask of the pan-Arctic LST dataset in GAC format has been applied to both datasets for comparison purposes.
}
\label{fig:Figure_04_EDLST}
\end{figure}

\begin{figure}
\centering
\includegraphics[width=\linewidth]{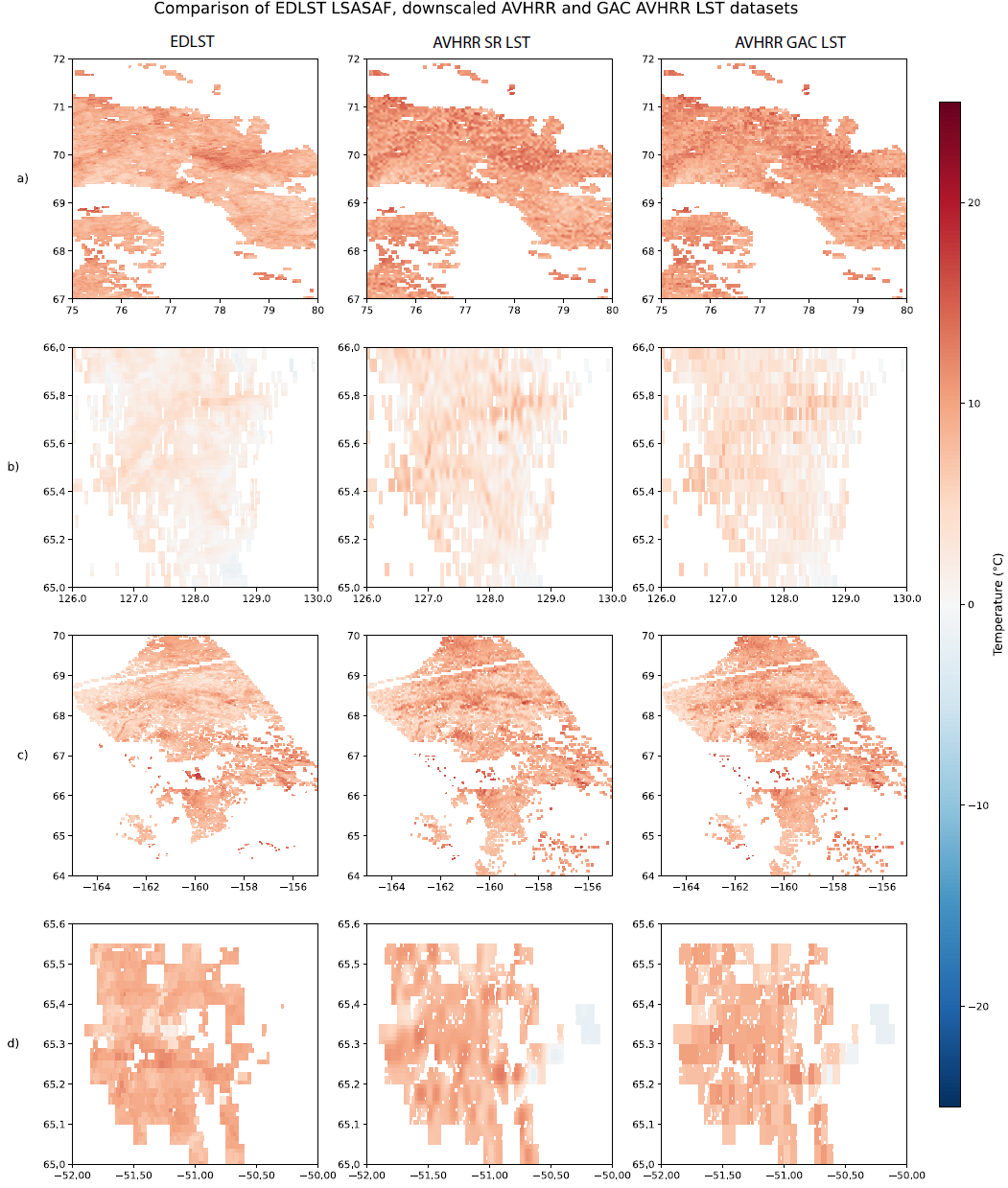}
\caption{Comparison of the EDLST product and the AVHRR SR LST dataset from the MetOp-B satellite during nighttime. a) On the 28.08.2020, in Northern Siberia, b) on the 09.08.2020 in East Siberia, c) on the 22.08.2020 in Alaska and d) on the 17.08.2020 in Greenland. The cloud mask of the pan-Arctic LST dataset in GAC format has been applied to both datasets for comparison purposes.
}
\label{fig:Figure_05_EDLST_night}
\end{figure}

\begin{figure}
\centering
\includegraphics[scale=0.5]{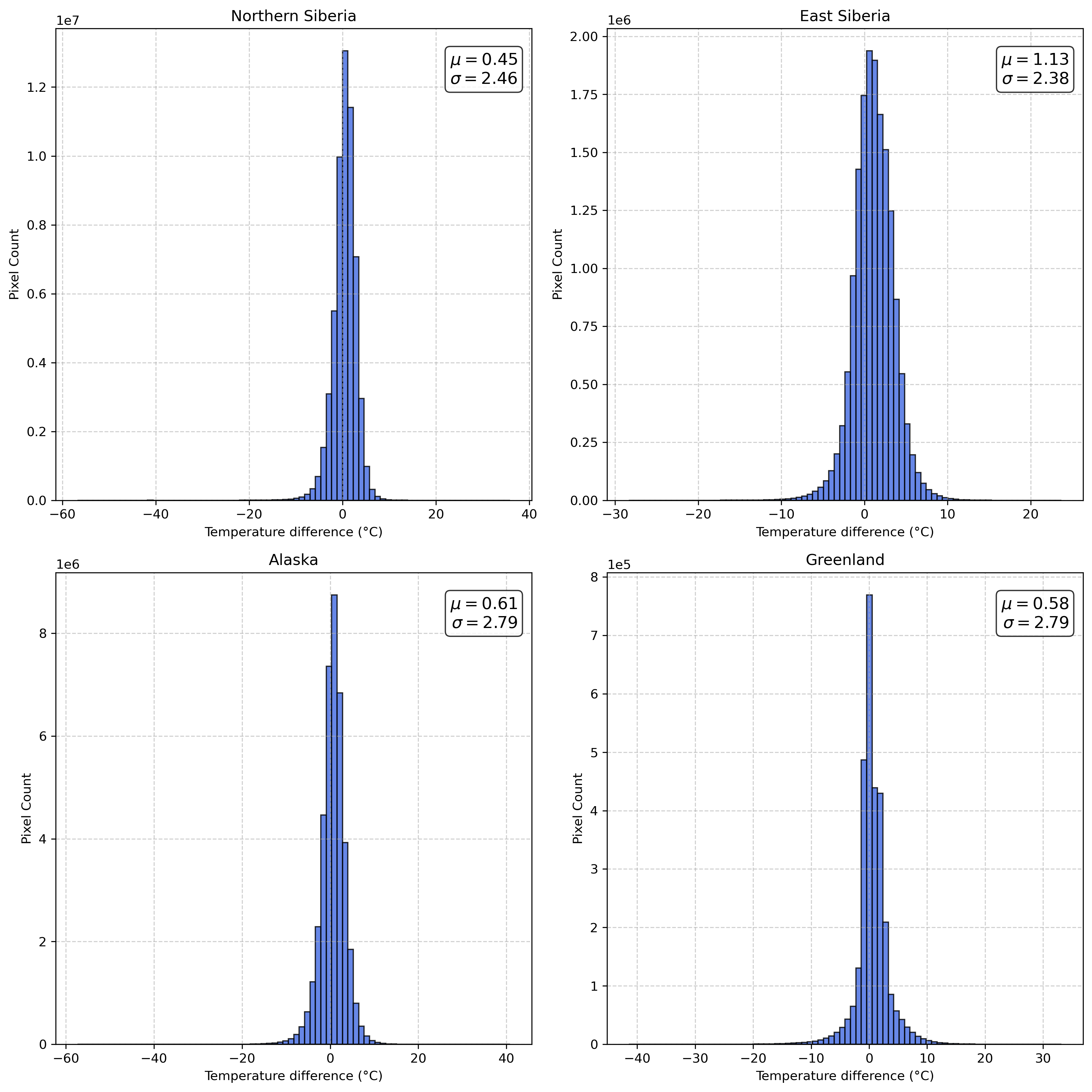}
\caption{Histograms of differences of Land SAF EDLST product minus the AVHRR SR LST from the MetOp-A satellite during daytime for four areas of interest (see Figure \ref{fig:Figure_04_EDLST}) for the entire year 2020.
}
\label{fig:hist}
\end{figure}

The EDLST product consists of daily composites of mean LST values and the cloud mask is obtained from the Nowcasting and Very Short Range Forecasting Satellite Application Facility (NWC) SAF. The pan-Arctic AVHRR LST GAC product has been generated by always selecting the pixel closest to the nadir and uses the cloud mask from the CLARA-A3 product \cite{Karlsson2023}. Figures  \ref{fig:Figure_04_EDLST} and\ref{fig:Figure_05_EDLST_night} show a comparison of the EDLST product and the AVHRR SR LST product for four areas of interest across the pan-Arctic during day and night respectively. The comparison uses the MetOp-A satellite at the daytime  and MetOp-B satellite at nighttime. The two LST products use different cloud masks: therefore in Figures \ref{fig:Figure_04_EDLST} and \ref{fig:Figure_05_EDLST_night}, the cloud mask used for the pan-Arctic product (CLARA-A3 cloud mask) is additionally applied to the already masked EDLST product. The EDLST product includes a water mask, therefore for the purpose of this comparison, the AVHRR SR LST dataset uses the ESA CCI land cover dataset for water masking. For the daytime comparison, both LST products show good agreement in all four regions (see Figure \ref{fig:Figure_04_EDLST}). The sharp temperature changes from the Greenland Ice Sheet to contiguous land is well represented in both datasets, and generally small structures are well visible in the downscaled dataset. In a few places, next to gaps due to clouds, the EDLST seems to be slightly cooler than the AVHRR SR LST e.g. a) and b) from Figure \ref{fig:Figure_04_EDLST}.
The nighttime comparison (Figure \ref{fig:Figure_05_EDLST_night}) presents small differences in temperature textures in Northern Siberia and the Greenland area, which could be attributed to differences in compositing as the GAC and the downscaled version are coherent. In addition, Figure \ref{fig:hist} shows the distribution of differences between the EDLST product and the MetOp-A AVHRR SR LST dataset for the same four areas of interest during daytime for the entire year 2020. The differences are computed as $EDLST - pan\_Arctic\_AVHRR\_LST(MetOpA)$.
The differences are centered around 0 \textdegree C and in three areas of interest. No cold or warm bias is visible from the distribution. East Siberia has a mean value of 1.13 \textdegree C, which indicates slightly warmer values in the EDLST product than in the AVHRR SR LST product.  Rare, very large deviations can be attributed to differences in cloud and water masking and to cloud contamination in one or the other dataset.

\clearpage

\section*{Usage Notes}

The presented 1 km AVHRR LST product, obtained from 40 years GAC LST with static high resolution guide images and a guided super-resolution algorithm from computer vision provides  long-term pan-Arctic LST observations at a hemispheric scale. It complements the existing LST time series derived from  MODIS, which is only available after 2000 and provides a valuable resource for assessing LST trends and dynamics over four decades. In addition, the proposed methodology and the dataset used to train the model can be reused for other similar use cases. The data are stored in NetCDF-4 format and follow the Climate-Forecast (CF) metadata conventions \cite{Hassell2017}. Python tools such as xarray or netcdf4 for example can be used to read the data.

The dataset contains only clear-sky observations and a conservative cloud mask has been used. Therefore, when studying LST trends or making comparisons, clear-sky bias might be present. As shown in Good et al. (2022)\cite{Good2022},  potential clear-sky bias may have be taken into account.

The land cover map used for the downscaling is static and does not account for the land cover changes over time. Furthermore, in areas that are permanently covered by ice, such as the Greenland ice sheet, the downscaling algorithm lacks guiding information as the land cover, the vegetation height, and the DEM will mostly be homogeneous. Users interested in downscaling  LST over ice sheets are encouraged to use other more fine-grained predictors. The training algorithm can be used seamlessly with different predictors.

\section*{Data Availability}

The pan-Arctic AVHRR LST dataset\cite{dupuis2025} is accessible from the data portal of the University of Bern (Boris Portal): \url{https://boris-portal.unibe.ch/handle/20.500.12422/222015}. The GAC and the downscaled LST product are packaged in different variables: 'LST' and 'LST\_GAC. The ESA CCI LST scenes and auxiliary data \cite{dupuis_2025_modis} used to train the DADA model are available from Zenodo: \url{https://doi.org/10.5281/zenodo.17341544}.

\section*{Code Availability}

The downscaling algorithm is written in the Python (v3.12.3) programming language, relying on the PyTorch framework, as well as TorchGeo for handling geospatial data. The Xarray and RioXarray packages have also been used for the analysis of the data in the NetCDF format.  The entire code used is freely available on GitHub (\url{https://github.com/soniajdupuis/Enhanced_pan_Arctic_LST}).

\pagebreak[2]

\end{document}